\documentclass{article}
\usepackage{spconf,amsmath,epsfig,soul}

\usepackage{amsmath,amsfonts,amssymb}
\usepackage{color}
\usepackage{tikz}

\usepackage{verbatim}
\usepackage{hyperref}
\usepackage{url}
\usepackage{placeins}
\usepackage{multirow,hhline}

\newcommand{\image}{\pgfuseimage}

\newcommand{\imPathC}{./images/P3/}

\pgfdeclareimage[width=0.3\columnwidth]{ref_3}{\imPathC ref.png}
\pgfdeclareimage[width=0.3\columnwidth]{BL_3}{\imPathC bl.png}
\pgfdeclareimage[width=0.3\columnwidth]{Dense_3}{\imPathC dense.png}
\pgfdeclareimage[width=0.3\columnwidth]{DenseL1_3}{\imPathC densel1.png}
\pgfdeclareimage[width=0.3\columnwidth]{Res_3}{\imPathC res.png}
\pgfdeclareimage[width=0.3\columnwidth]{ResL1_3}{\imPathC resl1.png}
\pgfdeclareimage[width=0.3\columnwidth]{cohe_3}{\imPathC cohe.png}
\pgfdeclareimage[width=0.3\columnwidth]{sar_3}{\imPathC sar.png}
\pgfdeclareimage[width=0.3\columnwidth]{inc_3}{\imPathC incid.png}

\newcommand{\imPathE}{./images/P5/}

\pgfdeclareimage[width=0.3\columnwidth]{ref_5}{\imPathE ref.png}
\pgfdeclareimage[width=0.3\columnwidth]{BL_5}{\imPathE bl.png}
\pgfdeclareimage[width=0.3\columnwidth]{Dense_5}{\imPathE dense.png}
\pgfdeclareimage[width=0.3\columnwidth]{DenseL1_5}{\imPathE densel1.png}
\pgfdeclareimage[width=0.3\columnwidth]{Res_5}{\imPathE res.png}
\pgfdeclareimage[width=0.3\columnwidth]{ResL1_5}{\imPathE resl1.png}
\pgfdeclareimage[width=0.3\columnwidth]{cohe_5}{\imPathE cohe.png}
\pgfdeclareimage[width=0.3\columnwidth]{sar_5}{\imPathE sar.png}
\pgfdeclareimage[width=0.3\columnwidth]{inc_5}{\imPathE incid.png}

\newcommand{\imPathG}{./images/P7/}

\pgfdeclareimage[width=0.3\columnwidth]{ref_7}{\imPathG ref.png}
\pgfdeclareimage[width=0.3\columnwidth]{BL_7}{\imPathG bl.png}
\pgfdeclareimage[width=0.3\columnwidth]{Dense_7}{\imPathG dense.png}
\pgfdeclareimage[width=0.3\columnwidth]{DenseL1_7}{\imPathG densel1.png}
\pgfdeclareimage[width=0.3\columnwidth]{Res_7}{\imPathG res.png}
\pgfdeclareimage[width=0.3\columnwidth]{ResL1_7}{\imPathG resl1.png}
\pgfdeclareimage[width=0.3\columnwidth]{cohe_7}{\imPathG cohe.png}
\pgfdeclareimage[width=0.3\columnwidth]{sar_7}{\imPathG sar.png}
\pgfdeclareimage[width=0.3\columnwidth]{inc_7}{\imPathG incid.png}

\definecolor{colTony}{rgb}{0.18, 0.2, 0.83}

\definecolor{colPeppe}{rgb}{0.75, 0.15, 0}

\newcommand{\rut}{\rule{0mm}{3mm}}

\usepackage{tikz,amsmath,amssymb,bm,color}
\usetikzlibrary{shapes,arrows}
\usetikzlibrary{calc,3d}
\usepackage{etoolbox}
\usepackage{multirow}

\newcommand{\labSty}[1]{{\scriptsize \sf \textbf{#1}}}
\newcommand{\Labcohe}{\labSty{Coherence}}
\newcommand{\Labsar}{\labSty{SAR}}
\newcommand{\Labincid}{\labSty{Incidence angle}}
\newcommand{\Labref}{\labSty{Reference}}
\newcommand{\Labbl}{\labSty{Baseline}}
\newcommand{\Labres}{\labSty{ResNet}}
\newcommand{\Labresl}{\labSty{ResNet (+$L_1$)}}
\newcommand{\Labdense}{\labSty{DenseNet}}
\newcommand{\Labdensel}{\labSty{DenseNet (+$L_1$)}}

\title{Deep Learning Solutions for {TanDEM-X}-based Forest Classification}
\name{Antonio Mazza$^{1}$, Francescopaolo Sica$^{2}$}
\address{
$^{1}$DIETI, University Federico II, Via Claudio 21, 80125 Naples, Italy\\
$^{2}$Microwaves and Radar Institute, DLR, M{\"{u}}nchener Stra{\ss}e 20, 82234 We{\ss}ling, Germany}

\begin{document}
%
\maketitle
\begin{abstract}
In the last few years, deep learning (DL) has been successfully and massively employed in computer vision for discriminative tasks, such as image classification or object detection.
This kind of problems are core to many remote sensing (RS) applications as well, though with domain-specific peculiarities. Therefore, there is a growing interest on the use of DL methods for RS tasks.
Here, we consider the forest/non-forest classification problem with TanDEM-X data, and test two state-of-the-art DL models, suitably adapting them to the specific task. Our experiments confirm the great potential of DL methods for RS applications.
\end{abstract}
\section{Introduction}
The monitoring of the state and health of forests is of primary importance for several reasons,
such as the prevention of floods and landslides, the reduction of CO$_2 $, or the preservation of biodiversity.
Thanks to the wide availability of optical, multispectral and synthetic aperture radar (SAR) data from a variety of sensors, 
such phenomena can be observed on a global scale provided that adequate processing tools are available.

Optical images carry rich discriminative information about vegetation and are widely employed. 
The Normalized Difference Vegetation Index (NDVI) is a notable example of a standard and simple vegetation indicator 
that is extracted through a straightforward combination of spectral bands \cite{Chakraborty2018}.
More specific indicators can also be derived from multispectral images, like the Enhanced Vegetation Index (EVI), more suited to discriminate canopy \cite{Pasquarella2018}.
However, the use of optical data is severely undermined by their dependence on the weather conditions,
which can be only partially mitigated through multitemporal processing and data fusion techniques \cite{Inglada2015, Scarpa2018, Errico2014}.
On the contrary, SAR data are almost weather insensitive and carry precious information related to ground geometry and electromagnetic propagation \cite{Gaetano2014}.
In \cite{Hagensieker2018} SAR images obtained in different bands are combined for land cover classification.
In \cite{Martone2018, Martone2018HighRes}, the TanDEM-X forest/non-forest map is generated from TanDEM-X bistatic interferometric images, 
by linking the presence of vegetation to the retrieved InSAR volume decorrelation.

In this work we focus on this latter problem, and experiment with deep learning solutions based on some state-of-the-art models.
Specifically, we define and train from scratch two DL architectures following the ResNet \cite{He2016} and the DenseNet \cite{Huang2017} models, respectively.
These are adapted to the problem at hand and to the available dataset, also through the definition of suitable loss functions.
Forest/non-forest classification maps obtained for a test area located in Pennsylvania confirm the great potential of DL approaches for RS classification tasks.
In Section \ref{sec:approach} we describe the approach and the details of the two proposed DL networks. 
Performance indicators, dataset, and experimental results are presented in Section \ref{sec:results}.
Finally, conclusions are drawn in Section \ref{sec:conclusions}.

\section{Proposed deep learning approach}
\label{sec:approach}

Deep learning models are characterized by an extremely large number of parameters to be trained, ranging from hundreds of thousands to billions, 
and organized in interconnected {\em layers} in order to generate a hierarchy of representations of the input.
Convolutional Neural Networks (CNNs) are a popular family of DL models, particularly suited to solving image processing problems.
In fact, under the assumptions of locality and shift-invariance, 
they adopt limited receptive fields and weight re-use, thereby ensuring a drastic reduction of the number of free parameters.
In this work, we consider two state-of-the-art CNN models, ResNet \cite{He2016} and DenseNet \cite{Huang2017},
which are particularly appealing as they can be reach a considerable depth avoiding vanishing gradient problems during training.
Both solutions are modular, allowing to build a variety of different architectures, 
from a simple cascade structure of an arbitrary number of layers to multipath architectures differing in the layer definition.
For both models, we describe here only the main functional aspects of interest for the present work, 
referring to the original papers \cite{He2016, Huang2017} for a thorough description of the network architecture.

In DenseNet, each layer is ``densely'' connected to all preceding ones. 
Therefore, the input of the $l$-th layer is obtained by concatenating the output features from all previous $l$-1 layers, not just the previous one.
This approach, with direct connections between each pair of layers, mitigates vanishing gradient and overfitting problems for large scale tasks.
In ResNet, instead, the training phase is shortened by using stages (one or a few consecutive layers)
whose output is the combination of the input (via skip connection) with the actual outcome of the trainable backbone.
Although functionally unnecessary, skip connections have proven to speed-up the training \cite{He2016, Scarpa2018a}

Here we propose for both models a cascade architecture with six convolutional layers with 3$\times$3 kernels interleaved by ReLU (Rectified Linear Unit) activation functions \cite{Krizhevsky2012}.
Moreover, in order to output a classification probability map,
an additional 1$\times$1 convolutional output layer with a sigmoid activation function completes the network.
The hyperparameters of the networks are summarized in Table \ref{tab:param}.
The 3-band input stack is formed by the SAR backscatter $\beta_0$, the interferometric coherence, and the local incidence angle, the latter obtained from the acquisition geometry and an external reference digital elevation model.

\begin{table*}
\centering
\setlength\tabcolsep{1.0pt}
{\footnotesize
\begin{tabular}{l@{\rule{-1pt}{0pt}}ccccccc} \hline
\rut               & ~~~ConvLayer 1~~~ & ~~~ConvLayer 2~~~  & ~~~ConvLayer 3~~~ & ~~~ConvLayer 4~~~  & ~~~ConvLayer 5~~~& ~~~ConvLayer 6~~~ &~~~ConvLayer 7~                \\ \hline

\rut Shape (ResNet)        & 64$\times$3$\times$3$\times$3 &   64$\times$64$\times$3$\times$3         &64$\times$64$\times$3$\times$3
			&   64$\times$64$\times$3$\times$3    &64$\times$64$\times$3$\times$3 & 64$\times$64$\times$3$\times$3     & 1$\times$64$\times$1$\times$1 \\
\rut Shape (DenseNet)        & 64$\times$3$\times$3$\times$3 &   64$\times$67$\times$3$\times$3         &64$\times$131$\times$3$\times$3 &   64$\times$195$\times$3$\times$3    &64$\times$259$\times$3$\times$3 & 64$\times$323$\times$3$\times$3     & 1$\times$64$\times$1$\times$1 \\
\rut Activation & ReLU & ReLU & ReLU & ReLU & ReLU & ReLU & sigmoid\\
\hline
\end{tabular}
}

\caption{CNNs' hyper-parameters. Shape: \# features $\times$ \# channels $\times$ 2D support. 
}
\label{tab:param}
\end{table*}

\subsection{Training}
In this work we exploit the same dataset used in \cite{Martone2018}, described in Section~\ref{sec:results}, 
including the ground-truth reference which is given in terms of density of forest in a squared area of 6$\times$6 meters.
In order to train the network we explored two different objective loss functions.
The former combines two losses commonly used for classification and segmentation, based on cross-entropy ($L_{bce}$) and on the Jaccard distance ($L_J$).
The latter includes also the $L_1$ norm, in order to reduce the absolute difference between the reference and the predicted density map.
In formulas, the cross-entropy loss reads as
\begin{equation}
    L_{bce} = - \frac{1}{N}\sum_n\left[ y_n\,\log{\left(\hat{y}_n\right)} + (1-y_n)\,\log{\left(1-\hat{y}_n\right)}\right],
\end{equation}
with $N$ being the number of pixels in a training batch, 
$y_n$ the class membership degree of pixel $n$ according to the ground-truth,\footnote{We assume the ground-truth density map as membership degree.} and $\hat{y}_n$ the membership estimated by the network.
The Jaccard distance loss, which aims to maximize the overlap between the two maps \cite{Csurka2013}, is defined as
\begin{equation}
    L_{J} = 1- \dfrac{\sum_n\left[y_n \cdot \hat{y}_n\right]}{\sum_n\left[y_n+\hat{y}_n - y_n \cdot \hat{y}_n\right]}.
\end{equation}
Finally, the $L_1$ norm is
\begin{equation}
    L_1 = \frac{1}{N} \sum_n \left| y_n-\hat{y}_n \right|.
\end{equation}

The minimization of the loss function is performed using the ADAM algorithm \cite{Kingma2014}, a gradient descend variant where the learning rate is updated at each iteration using estimates of low-order moments.

\section{Experimental results}
\label{sec:results}

The region of interest of the available dataset is located in Pennsylvania (USA).
We used 18707 tiles of 128$\times$128 pixels for training (90\%) and validation (10\%).
Tiles are grouped in mini-batches of 32 samples for the iterative optimization.
For each configuration the initial learning rate was set to $10^{-4}$ and the training was carried out from scratch for 20 epochs.\footnote{A pass on the whole training dataset.}
Five large images not used for training, of about 1800$\times$1450 pixels, were used to test the performance of the proposed methods.
These latter were chosen to be representative of the different environmental contexts.

Two performance indicators related to classification accuracy and segmentation accuracy are considered.
Following the methodology used in \cite{Martone2018}
we have chosen the accuray indicator ACC, a widespread quality index for binary classification problems, defined as
\begin{equation}
    \text{ACC} = \frac{\text{TP+TN}}{\text{TP+FP+TN+FN}},
\end{equation}
where TP, TN, FP, and FN count true positive, true negative, false positive, and false negative pixels, respectively.
In addition, in order to measure performance from the perspective of segmentation,
we also considered the Intersection-over-Union (IoU) indicator, which is the intersection between predicted and reference masks over their union.
For binary masks it is given by
\begin{equation}
    \text{IoU} = \frac{\text{TP}}{\text{TP+FP+FN}}.
\end{equation}
Both indicators fall between 0 (worst case) and 1 (ideal prediction).
Notice that the above definitions apply for binary images while the network is trained on probability values.
Therefore, to make them suited to our problem we decided to properly threshold both reference and predicted maps to get the needed binary masks.
To this end we followed the same criterion used in \cite{Martone2018},
maximizing the Pearson coefficient $\phi$ with respect to the threshold pair (one for prediction, one for reference).
This coefficient is defined as
\begin{equation}
    \phi = \frac{\text{TP}\cdot\text{TN} - \text{FP}\cdot\text{FN}}{\sqrt{\text{P}\cdot\text{RP}\cdot\text{RN}\cdot\text{N}}},
\end{equation}
where P[N] is the total number of positives[negatives] in the prediction map,
RP is the number of positives in the reference (RP=TP+FN) and, conversely, RN is the number of negatives in the reference (RN=FP+TN).
The Pearson coefficient is a sort of correlation coefficient which takes values between -1 and +1.
The threshold pair that corresponds to the maximum value of $\phi$ is the optimal choice according to this criterion (see \cite{Martone2018} for a deeper discussion).
In our implementation we used part of the validation set to find the optimal thresholds.

By doing so we eventually collect the performance indicators gathered in Table \ref{tab:iou} and Table \ref{tab:acc}, for IoU and ACC, respectively.
\begin{table}
\centering
\setlength\tabcolsep{1.0pt}
{\footnotesize
\begin{tabular}{l@{\rule{-1pt}{0pt}}ccccc} \hline
\rut               & ~Region 1~ & ~Region 2~  & ~Region 3~ & ~Region 4~  & ~Region 5~ \\ \hline
\rut Baseline \cite{Martone2018}       & 0.4540  &  0.4592 &   0.4245  &  0.6897 &   0.4644\\
\hline
\rut ResNet & 0.5960  &  0.6242 &   0.6915 &   0.8128 &   0.5686\\
\rut ResNet (+$L_1$) & 0.6013 &   \textbf{0.6362} &   0.7025 &   \textbf{0.8354}    &0.5885\\
\rut DenseNet &\textbf{0.6062} &   0.6354&    \textbf{0.7087}&    0.8306&    \textbf{0.5946}\\
\rut DenseNet (+$L_1$) & 0.5936  &  0.6105 &    0.6868    &0.8175&    0.5746\\
\hline
\end{tabular}
}
\caption{Intersection-over-Union comparison. +$L_1$ marks models trained using the full loss $L =L_{bce}+L_J+L_1$, other\-wise limited to the first two terms.}
\label{tab:iou}
\end{table}

\begin{table}
\centering
\setlength\tabcolsep{1.0pt}
{\footnotesize
\begin{tabular}{l@{\rule{-1pt}{0pt}}ccccc} \hline
\rut               & ~Region 1~ & ~Region 2~  & ~Region 3~ & ~Region 4~  & ~Region 5~ \\ \hline
\rut Baseline \cite{Martone2018}       & 0.6032 &   0.5964  &  0.6030    &0.7636&    0.8083\\
\hline
\rut ResNet & 0.8205 &    0.8290    &0.8769&    0.8766&    0.9014\\
\rut ResNet (+$L_1$) &0.8125 &    0.8259&    0.8771&    \textbf{0.8901}&    0.9023\\
\rut DenseNet &0.8231  &  \textbf{0.8309}&    \textbf{0.8841}&    0.8889 &   0.9097\\
\rut DenseNet (+$L_1$) & \textbf{0.8266}  & 0.8241 &   0.8788 &   0.8829 &   \textbf{0.9107}\\
\hline
\end{tabular}
}
\caption{Accuracy comparison.}
\label{tab:acc}
\end{table}
The numerical results speak clearly in favor of the proposed DL solutions, with DenseNet slightly outperforming ResNet, on average.
The use of the $L_1$ norm provides a negligible contribution, 
likely because the other two loss terms, based on cross-entropy and Jaccard distance, are directly related to ACC and IoU, respectively.
It has to be remarked that for a fair comparison with the proposed methods,
the baseline solution of \cite{Martone2018} was used without masking any class, contrarily to what is done in the original formulation.
Specifically, in \cite{Martone2018} city and water classes are excluded by means of available masks,
because forests, cities and water classes all exhibit a low volume correlation, the core feature proposed to classify forests.

For a further analysis of the performance of the proposed solutions we show some sample images to highlight merits and critical aspects of our proposal.
In Fig.\ref{fig:P5}, a case is shown where our proposals work fairly well.
In this case, the baseline method also provides results that are coherent with the reference, but rather noisy.
\newcommand{\DATE}{_5}
\begin{figure}[!ht]
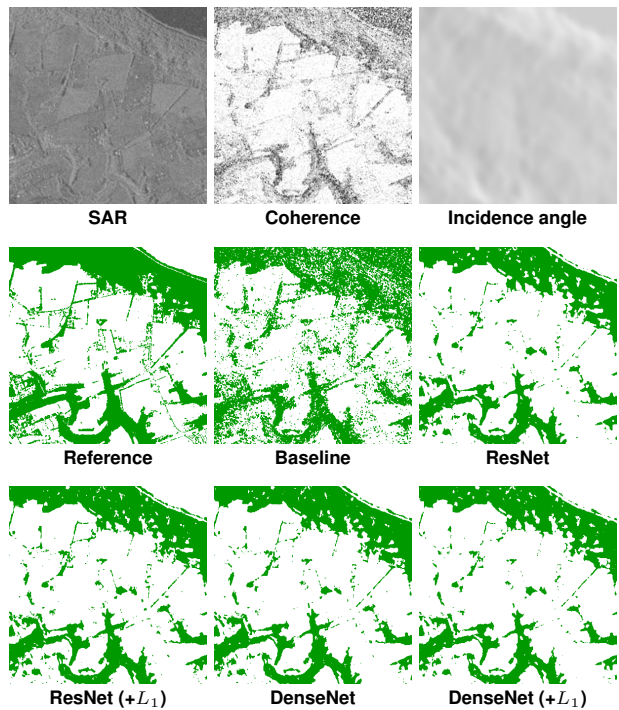

    \centering
    \begin{tabular}{ccc}
    	\image{sar\DATE} &\image{cohe\DATE}&  \image{inc\DATE}\\ [-1.5mm]
    	\Labsar& \Labcohe&  \Labincid\\[5pt]
    	\image{ref\DATE}& \image{BL\DATE} & \image{Res\DATE}\\ [-1.5mm]
    	\Labref& \Labbl& \Labres\\[5pt]
    	\image{ResL1\DATE}& \image{Dense\DATE} & \image{DenseL1\DATE}\\ [-1.5mm]
    	\Labresl& \Labdense& \Labdensel\\[5pt]
    	
    \end{tabular}
    \caption{The mask produced by deep learning approaches are clean compared to the baseline.}
    \label{fig:P5}
\end{figure}
In Fig.\ref{fig:P3} the occurrence of a limited number of false positive (on bridges) can be easily observed for all DL methods.
Moreover, some oversmoothing is also noticeable.
On the other side, the baseline method falls in a typical failure case, where it is unable to discriminate among forests, water and man-made areas.
\newcommand{\DATEA}{_3}
\begin{figure}[!ht]
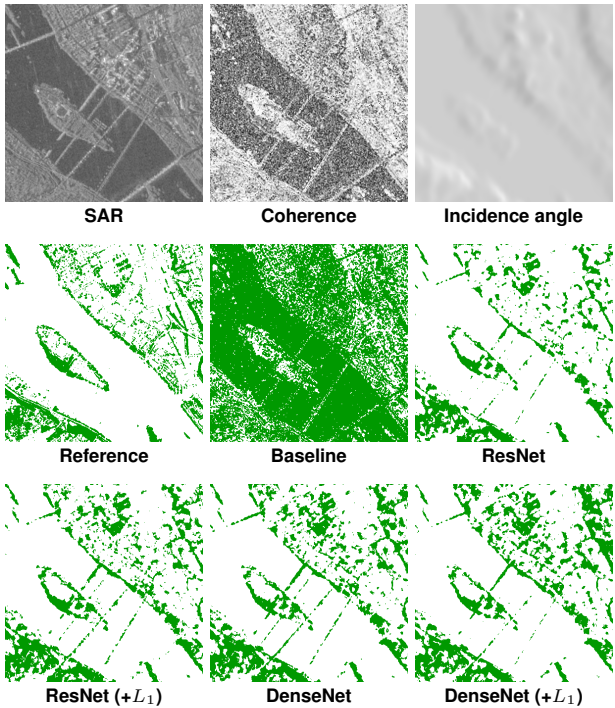

    \centering
    \begin{tabular}{ccc}
    	\image{sar\DATEA} &\image{cohe\DATEA}&  \image{inc\DATEA}\\ [-1.5mm]
    	\Labsar& \Labcohe&  \Labincid\\[5pt]
    	\image{ref\DATEA}& \image{BL\DATEA} & \image{Res\DATEA}\\ [-1.5mm]
    	\Labref& \Labbl& \Labres\\[5pt]
    	\image{ResL1\DATEA}& \image{Dense\DATEA} & \image{DenseL1\DATEA}\\ [-1.5mm]
    	\Labresl& \Labdense& \Labdensel\\[5pt]
    	
    \end{tabular}
    \caption{False positives and oversmoothing for DL solutions and faliure of the baseline.}
    \label{fig:P3}
\end{figure}
Finally, Fig.\ref{fig:P7} shows a detail where all methods present many false negatives.
However, the consistency between all predictions suggests that either a change in the scene occurred with respect to the reference, 
or radar data are unable to identify vegetated areas, in this case, due to a more complex backscattering mechanism.
\newcommand{\DATEB}{_7}
\begin{figure}[!ht]
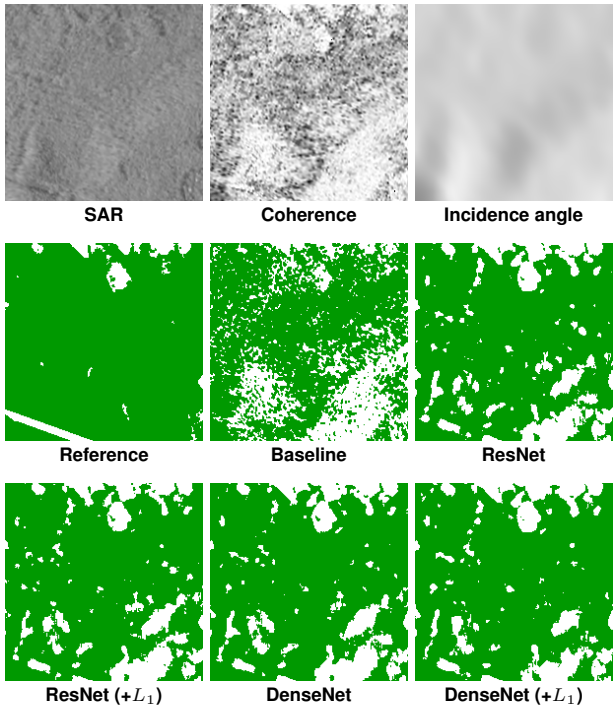

    \centering
    \begin{tabular}{ccc}
    	\image{sar\DATEB} &\image{cohe\DATEB}&  \image{inc\DATEB}\\ [-1.5mm]
    	\Labsar& \Labcohe&  \Labincid\\[5pt]
    	\image{ref\DATEB}& \image{BL\DATEB} & \image{Res\DATEB}\\ [-1.5mm]
    	\Labref& \Labbl& \Labres\\[5pt]
    	\image{ResL1\DATEB}& \image{Dense\DATEB} & \image{DenseL1\DATEB}\\ [-1.5mm]
    	\Labresl& \Labdense& \Labdensel\\[5pt]
    	
    \end{tabular}
    \caption{False negatives for all.}
    \label{fig:P7}
\end{figure}

\section{Conclusions}
\label{sec:conclusions}

In this work, we explored the use of deep learning methods for a forest/non-forest classification problem based on TanDEM-X data.
Despite the limited amount of labeled data available for training,
the proposed methods show very promising results, in terms of both objective numerical figures and subjective visual assessment.
More accurate results, especially in fine details preservation,
can be certainly obtained by using larger datasets for training, more sophisticated DL architectures, or additional hand-crafted features, which is the goal of our future work.

\small

\bibliographystyle{IEEEbib}
\bibliography{refs}

\end{document}